\pdfoutput=1

\documentclass[11pt]{article}

\usepackage[preprint]{acl}

\usepackage{times}
\usepackage{latexsym}

\usepackage[T1]{fontenc}

\usepackage[utf8]{inputenc}

\usepackage{microtype}

\usepackage{inconsolata}

\usepackage{graphicx}

\usepackage{multirow}
\usepackage{times}
\usepackage{soul}
\usepackage{url}
\usepackage[utf8]{inputenc}
\usepackage{caption}
\usepackage{graphicx}
\usepackage{amsmath}
\usepackage{amsthm}
\usepackage{booktabs}
\usepackage{algorithm}
\usepackage{algorithmic}
\usepackage[switch]{lineno}
\usepackage{multicol}
\usepackage{subcaption}
\usepackage{arydshln}
\usepackage[normalem]{ulem}
\usepackage{caption}
\usepackage{colortbl}
\usepackage{tcolorbox}
\usepackage{enumitem}
\usepackage{marvosym}
\newcommand{\nosection}[1]{\vspace{4.5pt}\noindent\textbf{#1}}

\title{Picking the Cream of the Crop:\\ Visual-Centric Data Selection with Collaborative Agents}

\author{
    \begin{tabular}{c}
  Zhenyu Liu$^1$, Yunxin Li$^1$, \textbf{Baotian Hu$^{1}$\textsuperscript{\Letter}\thanks{\textsuperscript{\Letter}Corresponding author.},} \\
  Wenhan Luo$^2$, Yaowei Wang$^1$, Min Zhang$^1$ \vspace{.5mm} \\
    \end{tabular}
    \\ \vspace{.5mm}
    \begin{tabular}{c}
    $^1$Harbin Institute of Technology (Shenzhen) \\
    $^2$Hong Kong University of Science and Technology \\
    \texttt{liuzhenyuhit@gmail.com}, \texttt{hubaotian@hit.edu.cn} \\
    \end{tabular}
     \\ \vspace{.5mm}
}

\makeatletter
\def\thanks#1{\protected@xdef\@thanks{\@thanks
        \protect\footnotetext{#1}}}
\makeatother

\begin{document}
\maketitle
\begin{abstract}
To improve Multimodal Large Language Models' (MLLMs) ability to process images and complex instructions, researchers predominantly curate large-scale visual instruction tuning datasets, which are either sourced from existing vision tasks or synthetically generated using LLMs and image descriptions. However, they often suffer from critical flaws, including misaligned instruction-image pairs and low-quality images. Such issues hinder training efficiency and limit performance improvements, as models waste resources on noisy or irrelevant data with minimal benefit to overall capability.
To address this issue, we propose a \textbf{Vi}sual-Centric \textbf{S}election approach via \textbf{A}gents Collaboration (ViSA), which centers on image quality assessment and image-instruction relevance evaluation. Specifically, our approach consists of 1) an image information quantification method via visual agents collaboration to select images with rich visual information, and 2) a visual-centric instruction quality assessment method to select high-quality instruction data related to high-quality images. Finally, we reorganize 80K instruction data from large open-source datasets.
Extensive experiments demonstrate that ViSA outperforms or is comparable to current state-of-the-art models on seven benchmarks, using only 2.5\% of the original data, highlighting the efficiency of our data selection approach. Moreover, we conduct ablation studies to validate the effectiveness of each component of our method. The code is available at \url{https://github.com/HITsz-TMG/ViSA}.

\end{abstract}

\section{Introduction}

Recent advancements in multimodal large language models (MLLMs) have led to significant progress in general visual understanding~\citep{DBLP:journals/corr/abs-2303-08774,DBLP:journals/corr/abs-2305-06500,DBLP:conf/nips/LiuLWL23a}. By integrating state-of-the-art large vision models (LVMs) with large language models (LLMs) and training on vast amounts of visual data, MLLMs have demonstrated the ability to comprehend complex images, achieving remarkable performance across a variety of visual tasks, including object recognition~\citep{DBLP:conf/cvpr/ZhangHMLLXQLLLG22}, visual question answering~\citep{DBLP:conf/icml/YuYLWL0WW24}, and image captioning~\citep{DBLP:conf/naacl/LevinboimTSS21}.
The training of MLLMs generally involves two main steps: first, pretraining on image-caption pairs to align the large vision model with the language model; second, instruction tuning on downstream visual tasks to enable the model to better understand complex images and perform sophisticated visual tasks.
To improve the multimodal understanding and task-solving abilities of MLLMs, existing approaches typically collect large-scale visual task data from a variety of sources for constructing instruction tuning datasets~\citep{DBLP:journals/corr/abs-2408-03326,DBLP:conf/nips/TongBWWIAYYMWPF24}.

\begin{figure}
    \centering
    \includegraphics[width=\columnwidth]{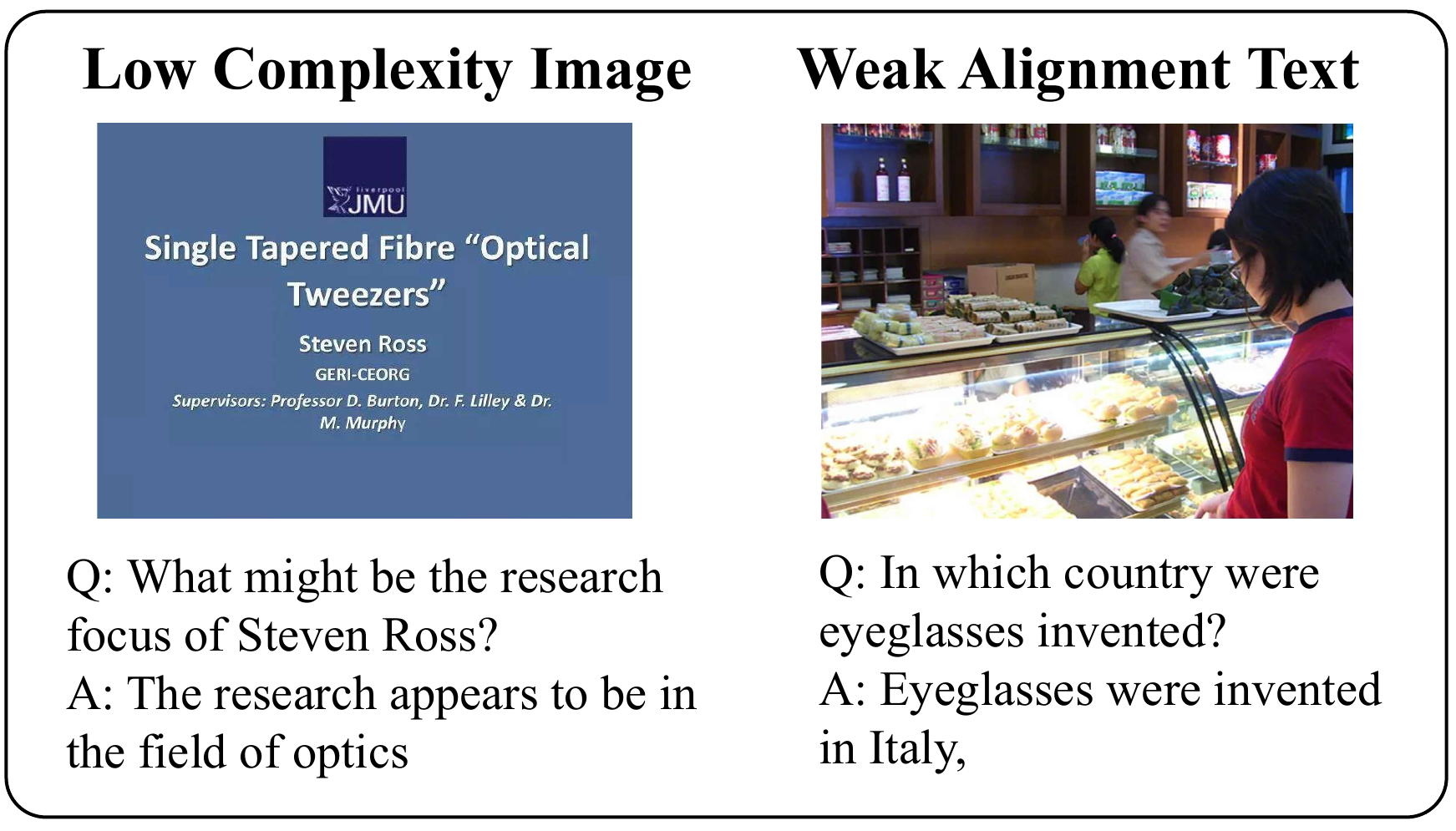}
    \caption{Two common issues in visual instruction tuning datasets. (Left)  Low-Complexity Image: The image lacks meaningful visual information. (Right) Low-Quality Text: The instruction is weakly aligned with the image.}  
    \vspace{-4mm}
    \label{fig:motivation}
\end{figure}

Specifically, existing works have primarily focused on collecting open-source training datasets from existing visual tasks~\citep{DBLP:journals/corr/abs-2408-03326,DBLP:conf/cvpr/LiuLLL24,DBLP:conf/nips/TongBWWIAYYMWPF24}, or synthetically generating various instructions using GPT-4 based on the collected images and description~\citep{DBLP:conf/nips/LiuLWL23a,DBLP:journals/corr/abs-2406-19736}. Such approaches enable the collection of large amounts of instructional data at a relatively low cost, without the need for extensive manual annotation. 
However, they often introduce significant noise into the training datasets. For example, images with low complexity or irrelevant instruction-image pairs can be problematic.  
As illustrated in Figure~\ref{fig:motivation}, the left subfigure shows an image of a presentation slide, which contains only a simple icon and basic presentation information. Such images provide little useful visual content or knowledge, making it difficult for models trained on such images to generalize to more complex real-world applications, where informative visuals are essential for learning robust visual representations. Another common issue is when the instruction is weakly aligned with the image, as shown in the right subfigure, where the instruction is a common-sense question that can be answered without referencing the image. This type of data fails to effectively link the visual model's perception with the textual knowledge stored in the language model, reducing the efficiency of multimodal learning.

To address this issue, we present a novel approach called \textbf{Vi}sual-Centric \textbf{S}election via \textbf{A}gents Collaboration (ViSA). This method leverages collaboration among multiple visual agents to assess the quality of visual data, selecting high-quality images and instruction pairs to enhance the training efficiency of MLLMs. The approach consists of two key components: 1) \textbf{Visual Information Quantification}, where agents collaborate to quantify visual elements and assess the richness of diverse image perspectives, selecting informative images; and 2) \textbf{Image-Centric Instruction Quality Quantification}, which focuses on selecting instructions tightly coupled with the images. This is achieved by leveraging multiple agents to calculate metrics such as prior token perplexity and image-text mutual information, enabling the evaluation of response quality and its relevance to the visual content. Additionally, to improve the collaboration across multiple visual agents, we introduce a Shapley value based on the Pearson correlation coefficient to weigh the reliability of each agent’s assessment.

To thoroughly assess the effectiveness of our method, we filter out 80K high-quality instruction data from the LLaVA-OneVision dataset~\cite{DBLP:journals/corr/abs-2408-03326}. We then conduct extensive experiments to evaluate the training effectiveness of the filtered data.
The results on seven visual benchmarks show that a model trained with less than 5\% of the data used by baselines achieves performance comparable to, or even surpassing, state-of-the-art models, highlighting the importance of high-quality data for understanding intricate images.
Furthermore, we observe significant improvements when continuing the training of a 72B model with a small set of high-information-density data, demonstrating that larger models can still benefit from informative images and well-aligned instructions. 
Overall, our agent collaboration-based data selection method effectively reduces noise in training datasets, leading to a significant increase in MLLM training efficiency.

Our main contributions can be summarized as follows:
\begin{itemize}[leftmargin=*]

\item We propose a novel multi-agent framework for quantifying and selecting high-quality visual instruction data. This framework evaluates both image informativeness and instruction relevance, leveraging a diverse set of visual agents to assess general visual elements as well as diverse image-specific features. To the best of our knowledge, this is the first work to explicitly introduce image informativeness evaluation in the context of visual data selection.

\item We introduce a new instruction quality quantification method based on prior token perplexity and image-text mutual information, which measure the response quality for selecting instructions tightly coupled with the images, respectively. We propose the Shapley value approach based on the Pearson correlation coefficient to effectively combine the evaluations from multiple agents.

\item We conduct extensive experiments and ablation studies, demonstrating that our approach significantly improves the training efficiency of MLLMs. Using only 80K selected samples, our method enables both 2B-scale and 7B-scale vision models to achieve performance on par with state-of-the-art models. Moreover, with a larger 72B-scale model, our high-quality selected data consistently enhances the model’s performance in visual understanding and reasoning tasks.

\end{itemize}

\section{Related Work}

\paragraph{Vision Large Models}: The development of Vision Large Models (VLMs) has led to significant advances in visual comprehension and reasoning~\citep{DBLP:conf/icml/0001LXH22,DBLP:conf/icml/0008LSH23,DBLP:conf/iclr/Zhu0SLE24,DBLP:conf/nips/LiuLWL23a,DBLP:conf/cvpr/LiuLLL24,DBLP:journals/corr/abs-2405-11273}. The training of multimodal large models (MLLMs) typically consists of two key stages. First, vision-language alignment is achieved by mapping the visual representations obtained from a pretrained Large Visual Model (LVMs)~\citep{DBLP:journals/corr/abs-2303-15389,DBLP:conf/icml/RadfordKHRGASAM21} into the LLM representation space through a learnable projection network~\citep{DBLP:conf/iclr/MerulloCEP23,DBLP:conf/icml/0008LSH23}. Subsequently, the aligned models are fine-tuned using instruction data to enhance their ability to process complex visual instructions~\citep{DBLP:journals/corr/abs-2312-14238,DBLP:journals/corr/abs-2308-12966,DBLP:journals/tmm/LiHCMXZ24,DBLP:conf/nips/Dai0LTZW0FH23}. Recently, there has been a growing trend in high-resolution MLLMs~\citep{DBLP:conf/eccv/GuoXYCNGCLH24,DBLP:journals/corr/abs-2407-07895,DBLP:journals/corr/abs-2308-12966,DBLP:journals/corr/abs-2409-12191}. Consequently, the ability to benefit from complex images has become increasingly important as informative images that can maximize these models' potential. Hence, our work aims to develop a quantitative assessment of image informativeness to select high-quality data.

\paragraph{Vision Instruction Datasets}: Instruction tuning aims to enhance a model's ability to understand and reason about images through diverse vision-language tasks, such as visual question answering~\citep{DBLP:conf/iccv/AntolALMBZP15}, image captioning~\citep{DBLP:conf/icml/RadfordKHRGASAM21}, and object detection~\citep{DBLP:journals/corr/abs-2408-00714}. MultiInstruct~\citep{DBLP:conf/acl/XuSH23} introduced the first human-annotated multimodal instruction tuning dataset, designed to enhance the zero-shot capabilities of pretrained VLMs. Building upon this foundation, subsequent studies have increasingly focused on curating diverse image datasets and harnessing the power of large multimodal models, such as GPT-4V, to automatically generate large-scale visual instruction data~\citep{DBLP:journals/corr/abs-2408-03326, DBLP:conf/acl/XuFSASJCWH24}. Several recent studies have explored different strategies to improve visual instruction tuning. Some works~\citep{DBLP:journals/corr/abs-2306-17107,DBLP:conf/aaai/HuXLLCT24} employ OCR tools and GPT-4 to generate instruction-following tasks that help VLMs better understand text within images. Meanwhile, recent works~\citep{DBLP:journals/corr/abs-2311-07574, DBLP:conf/eccv/ChenLDZHWZL24} directly prompt GPT-4V with images as inputs to generate visual instruction tuning data.
Unlike previous works that focus on instruction generation strategies, our approach emphasizes high-quality data selection by systematically assessing image informativeness and image-instruction relevance.

\section{Method}
\label{sec:method}

\begin{figure*}
    \centering
    \includegraphics[width=0.8\textwidth]{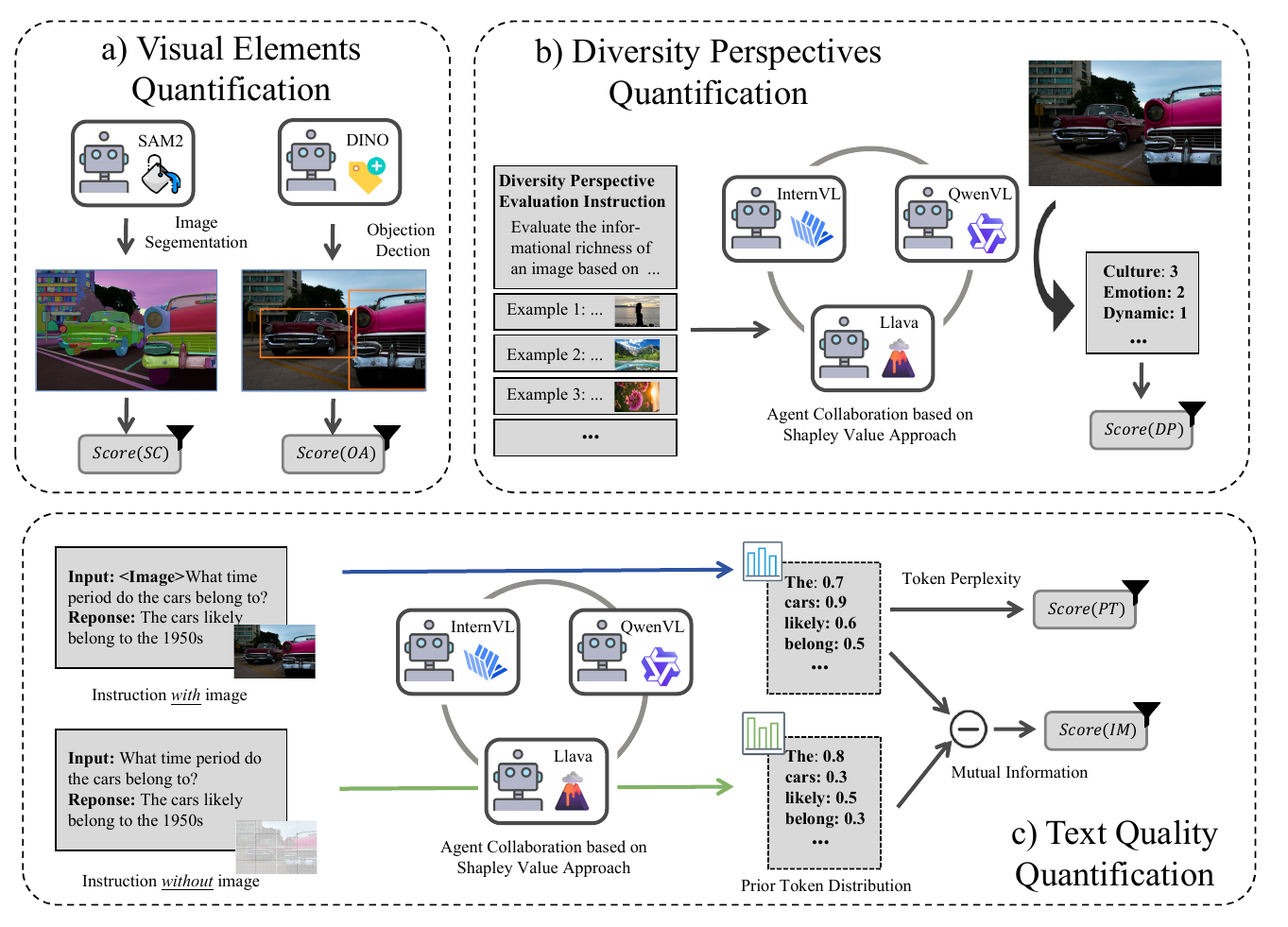}
    \caption{Overview of our agent collaboration for visual data selection. The $Score(SC)$ denotes the Segmentation Complexity Score. The $Score(OA)$ shows the Object Alignment Score. The $Score(DP)$ indicates the Diversity Perspective Score. The $Score(PT)$ denotes the Prior Token Perplexity Score.  The $Score(IM)$ denotes the Image-Text Mutual Information Score.}  
    \vspace{-4mm}
    \label{fig:framework}
\end{figure*}

In this section, we introduce our visual-centric selection via
agents collaboration. First, to identify informative images, we comprehensively consider various aspects of image information density assessment, including evaluation based on general visual elements~\S\ref{sec: Visual Elements Quantification} and assessment from other diversity visual information perspectives~\S\ref{sec: Diversity Perspectives Quantification}. Subsequently, we refine multimodal data selection on the textual level, incorporating text quality evaluation based on prior token perplexity and text relevance assessment based on mutual information~\S\ref{sec: Text Quality Quantification}. Finally, using our proposed method, we curated a high-quality instruction fine-tuning dataset~\S\ref{sec:Data Selection}.

\subsection{Visual Elements Quantification}
\label{sec: Visual Elements Quantification}

Evaluating image complexity from the perspective of visual elements is crucial for understanding image informativeness. However, existing methods primarily focus on object-level recognition or coarse-grained overall image assessment, lacking a fine-grain quantification of visual complexity. To address this, we assess image informativeness based on general visual elements (e.g., visual objects, graphical elements, characters, etc.), which are inherently linked to image complexity.
Specifically, we introduce two evaluation metrics: \textbf{Segmentation Complexity Score (SC Score)} and \textbf{Object Alignment Score (OA Score)}. The SC Score is designed to assess the richness of graphical elements and characters in an image, leveraging a visual segmentation method. In contrast, the OA Score evaluates the richness of objects in an image using the object detection approach based on a predefined category set.

Specifically, for the SC Score, we predefine 512 anchor points and employ the SAM2 model~\citep{DBLP:journals/corr/abs-2408-00714}, which segments target regions based on the provided points, to generate a set of segmentation boxes denoted as $M$. The SC Score is then computed using the following formula:
\begin{equation}
     \text{Score}(SC) = \text{count}(\text{IoU}(M, 0.75)).
\end{equation}

To evaluate the richness of objects, motivated by prior grounding work~\cite{DBLP:journals/corr/abs-2401-14159}, we first predefine a set of over 1800 common image object categories. Using the DINO model~\cite{DBLP:conf/eccv/LiuZRLZYJLYSZZ24}, which identifies the presence and location of predefined tags within images, we detect the visual objects corresponding to each category.
The OA Score is then computed by calculating the frequency of each category and normalizing it using the term frequency-inverse document frequency weight algorithm (TF-IDF), as shown in the following:
\begin{equation}
    \text{Score}(OA) = \sum_{i} \text{TF-IDF}(C_i),
\end{equation}
where $C_i$ is the predicted frequency of the $i$-th category, which emphasizes the importance of rare objects within the image set.

\subsection{Diversity Perspectives Quantification}
\label{sec: Diversity Perspectives Quantification}

We found that while the aforementioned visual element metrics provide a estimate of image informativeness with expectations, they are insufficient for fully capturing the richness of other types of image information. To address this limitation, we propose a multi-agent-based  \textbf{Diversity Perspective Score (DP Score)}, designed to evaluate images from multiple perspectives and dimensions, including events, emotions, culture, composition, and dynamics. These diverse perspectives offer a more comprehensive view of image informativeness, accounting for the varied origins and functional purposes of image creation. Specifically, we design a detailed evaluation guideline, complete with multiple examples, prompting various state-of-the-art visual agents to assess the image from each perspective and produce a final integrated score. 
For the specific evaluation guideline used, please refer to Appendix~\ref{sec:prompt appendix}.

To integrate the final score of each agent, we use a Shapley value approach~\citep{DBLP:conf/nips/LundbergL17} based on the Pearson correlation coefficient to compute the weight of each agent, motivated by the idea of determining model weights based on score consistency. Specifically, the weight $W_j$ of each agent is calculated as follows:
\begin{equation}
\begin{split}
    W_j = \sum_{S \subseteq N \setminus \{j\}} \omega(S) \Delta \rho(S, j), \\
    \omega(S) = \frac{|S|! (|N| - |S| -1)!}{|N|!}, \\
    \Delta \rho(S, j) = \rho(S \cup \{j\}) - \rho(S)
\end{split}
\end{equation}
where $N$ represents the set of all agents, and $S$ is any subset of agents excluding the $j$-th agent. $\text{Pearsonr}(S)$ is the Pearson correlation coefficient of the score distribution over the agents set $S$, measuring the linear relationship between the scores. The overall diversity perspective score for each image is then computed as the weighted average of the diversity perspective scores $E_j$ from multiple agents.
\begin{equation}
    \text{Score}(DP) = \sum^{j}_{|N|} W_j E_j.
\end{equation}

\subsection{Text Quality Quantification}
\label{sec: Text Quality Quantification}

To fully capture the visual and textual aspects of image informativeness, the quality of the associated text is also important. In this regard, we propose a multi-agent-based  text quality quantification method, which includes two key metrics: the  \textbf{Prior Token Perplexity Score (PT Score)}, which measures the quality of the textual content, and the \textbf{Image-Text Mutual Information Score (IM Score)}, which assesses the degree of association between the text and the image.
Similar to the approach in Section~\ref{sec: Visual Elements Quantification}, we utilize three different agents to compute the PT Score and the IM Score separately, and then aggregate these scores using the Shapley value approach to obtain their comprehensive quality score.

Specifically, the PT Score is calculated by evaluating the perplexity of the response, which is defined as follows. Due to the significant variability in the length distribution of the data instructions, we adopt a more length robust approach to prior token perplexity~\cite{DBLP:conf/emnlp/LiuLHZCHZ24}:
\begin{equation}
    P_j = \exp \left( -\frac{1}{N} \sum_{i=1}^{K} \log P(w_i | w_{<i}) \right),
\end{equation}
\begin{equation*}
   \text{Score}(TP) = \sum^j_{N} \sum_{S \subseteq N \setminus \{j\}} \omega(S) \Delta \rho(S, j) P_j,
\end{equation*}
where $K$ denotes the number of prior tokens. This score reflects how predictable or coherent the sequence of tokens is based on preceding words.

Next, we compute the IM Score, which measures the mutual dependence between the image and the associated text, formulated as following:
\begin{equation}
    I_j = H(\text{Text}) - H(\text{Text} | \text{Image}),
\end{equation}
\begin{equation*}
   \text{Score}(IM) = \sum^j_{N} \sum_{S \subseteq N \setminus \{j\}} \omega(S) \Delta \rho(S, j) I_j ,
\end{equation*}
where $H(\text{Text})$ is the entropy of the text without the image input, and $H(\text{Text} | \text{Image})$ is the conditional entropy of the text given the image. A higher mutual information value indicates a stronger association between the text and the image.

\subsection{Data Selection}
\label{sec:Data Selection}

To evaluate the effectiveness of our data selection approach, random sample 175K instruction data points from the open-source LLaVA-OneVision dataset~\citep{DBLP:journals/corr/abs-2408-03326}. We then applied the SC score, OA score, and DP score to filter out 15\%, 20\%, and 13\% of the data, respectively, resulting in 100K high-quality image samples. Subsequently, we further refined the dataset by applying the TP score and IM score to select 10\% of the data, yielding a final set of 80K high-quality instruction fine-tuning samples.
\section{Experiment}

\subsection{Evaluation}

To evaluate the performance of our method, we introduce three well-established image understanding benchmarks, including VQAv2~\citep{DBLP:conf/iccv/AntolALMBZP15}, OKVQA~\citep{DBLP:conf/cvpr/MarinoRFM19}, TextVQA~\citep{DBLP:conf/cvpr/SinghNSJCBPR19}, along with four cutting-edge datasets designed for complex visual comprehension and reasoning, including MMBench~\citep{DBLP:conf/eccv/LiuDZLZZYWHLCL24}, MME-RealWorld~\citep{DBLP:journals/corr/abs-2408-13257}, SEED-Bench~\citep{DBLP:journals/corr/abs-2307-16125}, MMMU~\citep{DBLP:conf/cvpr/YueNZ0LZSJRSWYY24}. We provide more detail in the Appendix~\ref{sec:evaluation_appendix}.

\subsection{Baselines}

We compare our approach against SOTA MLLMs, including Qwen2-VL~\citep{DBLP:journals/corr/abs-2409-12191}, Llava-OneVision~\citep{DBLP:journals/corr/abs-2408-03326}, Llama-3.2-Vision~\citep{DBLP:journals/corr/abs-2407-21783}, as well as InternVL2 and InternVL2.5~\citep{DBLP:journals/corr/abs-2312-14238}. all of which have been trained on massive-scale datasets with extensive computational resources.
To comprehensively evaluate the effectiveness of our method across different model scales, we conduct experiments using models of three representative parameter sizes: small-scale (2B parameters), medium-scale (7B–11B), and large-scale (70B–90B). 
Due to the computational constraints, we employ the INT4 quantized versions of all large-scale models. 

\begin{table*}[ht!]
\centering
\small
{
\setlength{\tabcolsep}{4pt}
	\begin{tabular}{l|c|ccccccc}
		\toprule
        \textbf{Models}                 &\textbf{Data Scale}      &\multicolumn{1}{c}{OKVQA}    &\multicolumn{1}{c}{VQAv2} &\multicolumn{1}{c}{TextVQA} &\multicolumn{1}{c}{MMBench} &\multicolumn{1}{c}{MME-RW} &\multicolumn{1}{c}{SEED} &\multicolumn{1}{c}{MMMU} \\ 
        \cline{1-9}
        InternVL2-2B                    &>5M        &41.2	          &75.9	          &72.0           &70.9	          &37.8	          &71.8	          &33.0     \\
        InternVL2.5-2B                  &>8M        &39.2	          &\textbf{77.8}  &71.9           &\textbf{79.0}  &34.6	          &\textbf{72.8}  &40.3       \\
        \cdashline{1-9}
        Qwen2-VL-2B                     &>5M        &44.3	          &71.5	          &77.6           &74.6	          &31.2	          &68.2	          &33.6         \\
        Qwen2-VL-2B-AC (ours)           &80K        &\textbf{48.4}	  &73.3	          &\textbf{81.4}  &73.0	          &\textbf{39.1}  &69.4	          &\textbf{42.2}         \\
        \hline\hline 
        Llama3.2-vision-11B             &>3M        &23.6	          &72.2	          &56.5	          &65.8           &38.5           &72.7           &48.1       \\
        Llava-Onevision-7B              &>1.6M      &\textbf{61.6}	  &82.5           &80.3	          &80.9           &48.6           &75.4         &47.9       \\
        InternVL2-8B                    &>5M        &46.4	          &79.2	          &77.9	          &79.4           &37.4          &75.4           &51.2       \\
        InternVL2.5-8B                  &>8M        &54.4	          &80.7	          &80.9           &\textbf{82.5}  &46.4          &\textbf{76.8}   &\textbf{56.2}       \\
        \cdashline{1-9}
        Qwen2-VL-7B                     &>5M        &58.0	          &83.0	          &\textbf{85.4}  &81.0           &48.3           &75.0   &50.6       \\
        Qwen2-VL-7B-AC (ours)           &80K        &54.2	          &\textbf{83.6}  &84.3	          &80.7           &\textbf{49.4}  &75.2   &49.9         \\
        \hline\hline   
        Llama3.2-vision-90B             &>3M        &46.7	          &73.1	          &71.0           &84.3           &45.4	          &75.8	      &38.2           \\
        Llava-Onevision-72B             &>1.6M      &\textbf{66.3}	  &84.8	         &82.6	         &86.4	          &48.8	          &77.8	      &50.6       \\
        InternVL2-76B                   &>5M        &60.9	          &84.6	          &87.1	          &87.3	          &\textbf{49.1}  &76.5	      &52.2       \\
        InternVL2.5-78B                 &>8M        &57.4	          &85.2	          &86.3	          &\textbf{91.9}  &46.4	          &77.0       &\textbf{62.5}         \\
        \cdashline{1-9}
        Qwen2-VL-72B                    &>5M        &52.9	          &83.2	          &90.8	          &90.6	          &40.8	          &78.3	          &59.1       \\
        Qwen2-VL-72B + AC-lora (ours)   &80K        &58.0             &\textbf{90.5}   &\textbf{91.3}   &89.8	      &47.4	          &\textbf{79.8}  &60.6       \\
        \hline
	\end{tabular}
 }
    \caption{Main result of visual understanding task. The "Data Scale" refers to the instruction data scale reported in each respective paper. Models with more than 72B parameters use InT4 quantification. The best results in each task are highlighted in \textbf{bold}.}
    \label{tab:main result}
    \vspace{-0.1cm}
\end{table*}

\begin{table*}[ht!]
\centering
\small
{
\setlength{\tabcolsep}{5pt}
	\begin{tabular}{l|l|ccccccc}
		\toprule
        \textbf{Models}                          &\textbf{Data Scale}      &\multicolumn{1}{c}{OKVQA}    &\multicolumn{1}{c}{VQAv2} &\multicolumn{1}{c}{TextVQA} &\multicolumn{1}{c}{MMBench} &\multicolumn{1}{c}{MME-RW} &\multicolumn{1}{c}{SEED} &\multicolumn{1}{c}{MMMU} \\ 
        \cline{1-9}
        Qwen2-VL-2B-AC                          &80K    &48.4	&73.3	&81.4   &73.0	&39.1	  &69.4	     &42.2         \\  
        \quad - Instruction Quality               &100K   &43.5	&72.2	&75.8   &70.8	&36.0     &68.1	     &38.3         \\  
        \quad - Diversity Perspective     &115K   &42.4	&72.6	&74.9	&69.7	&35.4	  &67.7	     &37.5         \\ 
        \quad - Visual Element            &175K   &41.0   &70.3   &72.1   &68.5	&33.9	  &67.1	     &35.8         \\   
        \hline\hline 
        Qwen2-VL-7B-AC                          &80K    &54.2	&83.6   &84.3	&80.7   &49.4     &75.2      &49.9         \\   
        \quad - Instruction Quality              &100K   &50.4   &81.9   &82.4   &77.3   &47.9     &73.2      &48.1         \\   
        \quad - Diversity Perspective     &115K   &49.3   &81.7   &81,9   &76.5   &47.0     &72.8      &48.2         \\   
        \quad - Visual Element            &175K   &47.2   &79.3   &79.1   &75.7   &45.3     &71.8      &44.1          \\  
        \hline
	\end{tabular}
 }
    \caption{Ablation study of small-scale and middle-scale on visual understanding task. The "Data Scale" refers to the training data size before filtering.}
    \label{tab:ablation result}
    \vspace{-0.4cm}
\end{table*}

\subsection{Implementation Detail}

We conduct full fine-tuning on Qwen2-VL-2B and Qwen2-VL-7B as our base models. Due to computational constraints, we adopt Qwen2-VL-72B-instruction as the base model for large-scale experiments and apply LoRA for efficient continued fine-tuning.
All models are optimized using AdamW~\cite{DBLP:conf/iclr/LoshchilovH19} with a learning rate of 1e-5 for full fine-tuning and 1e-4 for LoRA-based fine-tuning, along with a weight decay of 0.01. We employ a linear warmup strategy with a 0.1 warmup ratio and set the LoRA rank to 16. The maximum sequence length is 2048 tokens, and all models are trained and evaluated using the BFloat16 floating-point format.
For training efficiency, we use a batch size of 32 with four-step gradient accumulation on a single GPU. Additionally, we follow the original chat template of Qwen2-VL for model interactions. We used beam search with the beam size set to 4 for generation. We run all the experiments on the NVIDIA H800 GPU. 

\subsection{Main Result}

In this section, we explore the two following research questions: (1) Q1: How does our data compare with current mainstream training datasets in terms of efficiency? and (2) Q2: Can a small amount of high-quality data continuously improve a well-trained, large-scale model?

\nosection{Q1: A small amount of high-quality data significantly efficiency improves the instruction-following ability of pretrained MLLMs.}
As shown in Table~\ref{tab:main result}, with less than 5\% of the baseline training data, our method achieves comparable or superior performance across all datasets. Specifically, our Qwen2-VL-2B-AC model achieves an average score of 61.0\% on 7 benchmarks, outperforming Qwen2-VL-2B (57.3\%), InternVL2.5-2B (59.4\%), and InternVL2-2B (57.5\%). For middle-scale models, our Qwen2-VL-7B-AC model reaches an average score of 68.2\%, surpassing Llama3.2-vision-11B (53.9\%), InternVL2-8B (63.8\%), and matching the performance of Llava-Onevision-7B (68.1\%), Intern2.5-8B (68.3\%), and Qwen2-VL-7B (68.7\%).
These results suggest that the key factor for training pretrained MLLMs is not solely the amount of data, but rather the quality of the data. A small amount of high-quality data can significantly boost the multimodal instruction-following capability of pretrained MLLMs, enabling the model to activate the visual knowledge learned during pretraining and perform a variety of visual tasks.
Moreover, we observe that on the MME-RealWorld dataset, a complex real-world visual benchmarks, our model consistently outperforms the baseline, highlighting the crucial role of complex visual data in real-world applications.
However, we also see that compared to the small-scale model, the advantages of high-quality data are less pronounced in the middle-scale model. This may be because larger models require more high-quality data to fully fine-tune the deeper parameters, a finding also shown in the work of training scaling law~\citep{DBLP:journals/corr/abs-2001-08361}.

\nosection{Q2: A small amount of high-quality data can continuously boost large-scale MLLMs.} 
As shown in Table~\ref{tab:main result}, continued training on the Qwen2-VL-72B model leads to noticeable improvements in its image understanding capabilities. The Qwen2-VL-72B-AC model achieves 73.9\% on 7 datasets, surpassing other baseline models. Specifically, compared to the original Qwen2-VL-72B, our fine-tuned model shows significant gains on the two datasets that focus on complex image perception and understanding: OKVQA and MME-RealWorld (4.1\% and 6.6\% improvement, separately). These results suggest that even large models trained on vast amounts of data can still be hungry for high-quality data, and a small amount of high-quality data can rapidly boost their performance. Moreover, this also highlights the importance of high-quality data for effective continued instruction tuning in very large models.

\section{Ablation Study}

In this section, we conduct ablation experiments to assess the effectiveness of the proposed filtering methods, as presented in Table~\ref{tab:ablation result}. Our findings reveal that even when the total amount of data is increased by removing the specific filter, the performance quickly degrades. This suggests that fine-tuning with a small amount of high-quality data is significantly more efficient than using large amounts of noisy data. Our method effectively filters out noisy data, thereby improving the fine-tuning efficiency of the model.
Notably, the performance shows the most significant decline when the 
instruction quality filter is removed, with average performance drops of 3.1\% and 2.5\% for the small and middle-scale models, respectively. This confirms our observation that weakly aligned instructions in the current visual instruction data have a detrimental effect. Poorly aligned instruction data negatively impacts the perception and knowledge alignment between the visual encoder and the large language model, ultimately reducing the overall performance.
Moreover, the removal of the visual element filter also leads to a noticeable decrease in performance (1.6\% and 2.1\% for the small and middle-scale models, respectively). This emphasizes the critical role of image quality in data selection; low-quality or noisy images hinder the efficiency of visual representation learning by the vision encoder.




\begin{figure*}[ht]  
    \centering
    
    \begin{subfigure}[b]{0.32\textwidth}
        \centering
        \includegraphics[width=\linewidth, trim=50 30 50 30, clip]{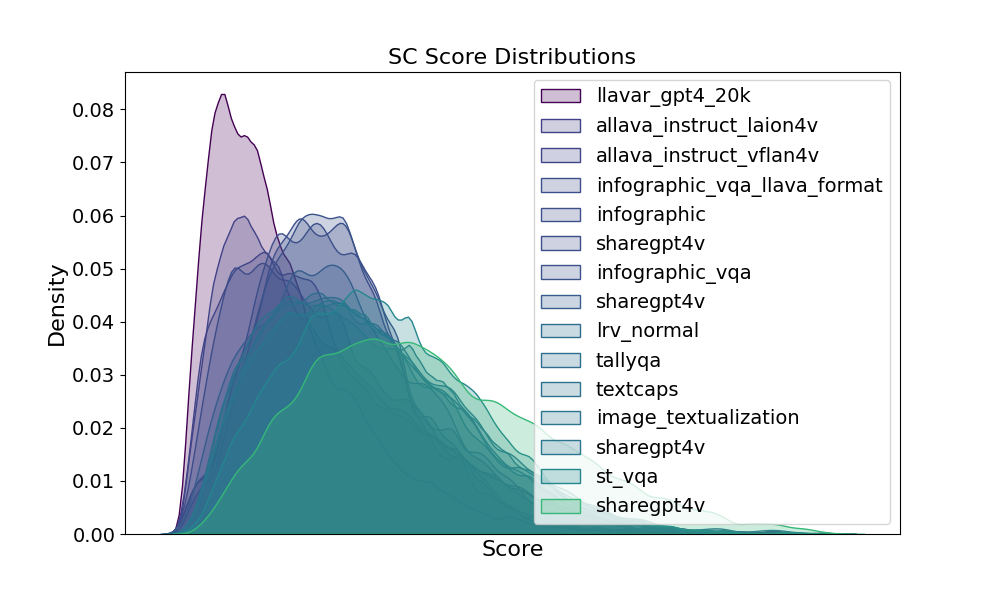}
        \caption{}
    \end{subfigure}
    \hfill
    \begin{subfigure}[b]{0.32\textwidth}
        \centering
        \includegraphics[width=\linewidth, trim=50 30 50 30, clip]{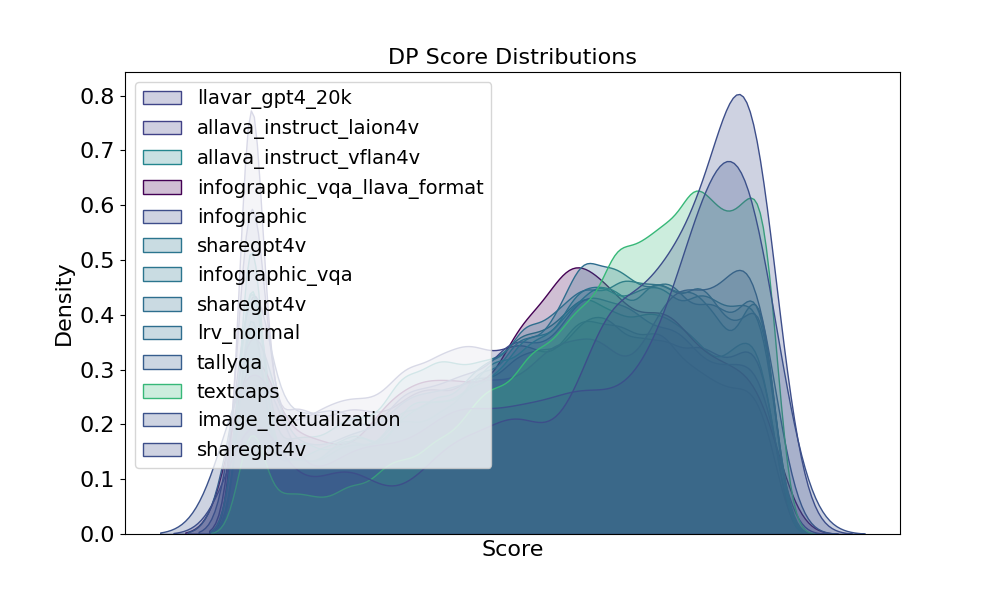}
        \caption{}
    \end{subfigure}
    \hfill
    \begin{subfigure}[b]{0.32\textwidth}
        \centering
        \includegraphics[width=\linewidth, trim=50 30 50 30, clip]{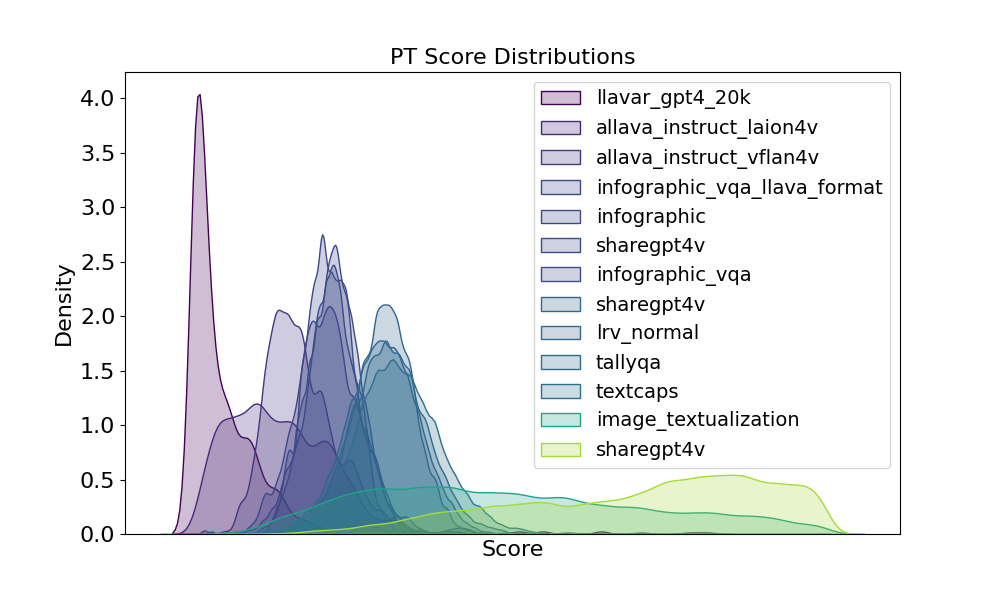}
        \caption{}
    \end{subfigure}
    
    \caption{Score distributions across different datasets: (a) segmentation complexity, (b) diversity perspectives and (c) prior token perplexity.}
    \label{fig:score_distributions}
    \vspace{-4mm}
\end{figure*}

\begin{figure}[ht]
    \centering
    \includegraphics[width=\linewidth]{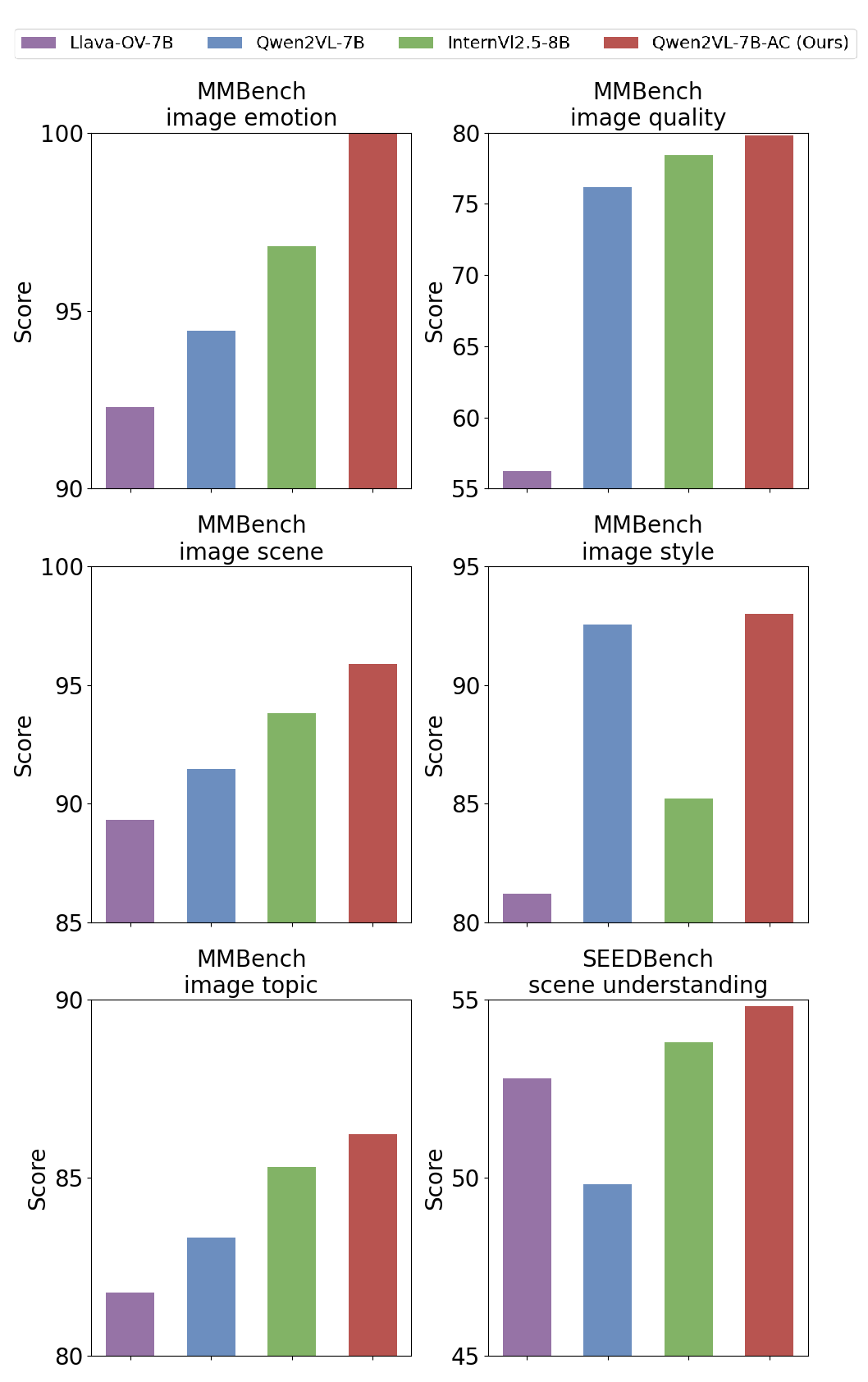}
    \caption{Results of the medium-scale model on six curated complex image understanding and perception datasets.}  
    \vspace{-4mm}
    \label{fig:complex understanding}
\end{figure}

\section{Complex Visual Understanding}

To fully explore the role of high-quality data in complex image perception and understanding, we selected six sub-tasks of image  comprehension from the practical related benchmark, including sentiment classification, quality assessment, scene perception, and style classification from MMBench, as well as scene understanding from SEED Bench and visual perception from MME-RealWorld. We report the results of our method on a 7B model setup, as shown in Figure~\ref{fig:complex understanding}. 

Our findings reveal that, unlike the similar overall performance with baselines in the main results, our Qwen2-VL-7B-AC model consistently outperforms the baseline across all complex image understanding tasks. This demonstrates that informative image data is crucial for improving a model's ability to perceive and understand complex images. While recent works~\citep{DBLP:conf/cvpr/LiYLMZYSLB24,DBLP:conf/eccv/GuoXYCNGCLH24} in visual model architectures often focus on increasing image resolution to enhance complex image understanding, higher resolution images do not necessarily equate to greater complexity. For example, compared to a high-resolution document scan with more content, a lower-resolution yet content-rich street scene image proves to be far more beneficial for a model’s visual representation learning. Images containing dense visual elements effectively activate the model’s ability to recognize and learn the relationships between these elements, thus bridging the gap between the visual model's low-level visual perception and the world knowledge stored in the pretrained language model.

\section{Quantification Distribution Analysis}

In this section, we present a visualization and detailed analysis of the proposed image informativeness metrics, focusing on three metrics, with the additional metrics provided in Appendix~\ref{sec:distrubution appendix}.

\paragraph{Segmentation complexity score distribution}
As shown in Figure~\ref{fig:score_distributions}, we observed that the distribution of the segmentation complexity score varies across different datasets. For example, the score of the ``llava\_gpt4\_20k'' dataset is relatively low, as it typically contains text-image pairs with limited visual content, resulting in fewer visual elements within the dataset. In contrast, the ``sharegpt4v'' dataset primarily includes scene images from real-life scenarios, which are generally richer in visual elements, leading to higher SC scores. This difference indicates that the ``sharegpt4v'' dataset contains more complex images.
This distribution demonstrates that our proposed segmentation complexity score effectively reflects the number of visual elements in the data, indirectly capturing the complexity of the images.

\paragraph{Diversity perspectives score distribution}
Unlike the distribution of other scores, the diversity perspectives score exhibits a notably extreme trend, with the visual agent tending to assign either the highest or the lowest ratings. Our analysis suggests that this is primarily due to the multifaceted nature of the image evaluation process, where images are assessed with multiple dimensions. The diversity in the types and sources of images further complicates the task of providing a comprehensive score. Additionally, we intentionally designed the system to prevent the agent from assigning low scores based on a few weak dimensions of an image. For example, we cannot devalue the quality of a close-up photograph of still life simply due to the lacking of visual events. As a result, the agent tends to classify images as either extremely poor or exceedingly rich in terms of visual content.
This underscores the substantial challenge in evaluating images across multiple dimensions of quality. The diverse characteristics of images make it difficult to achieve a balanced and nuanced assessment, which remains an open issue in multimodal evaluation tasks.

\paragraph{Prior token perplexity score distribution}
For analytical convenience, we apply the negative transformation to the prior token perplexity scores when visualization. Upon examining the distribution of the prior token perplexity scores, we observe significant variability across datasets. This can likely be attributed to the greater disparity in the response text across different tasks. While higher scores (shifted towards the right) indicate that the token probability distribution is more certain. However, excessively high scores may also reflect overly simplistic responses. Therefore, when setting the threshold for filtering prior token perplexity scores, we exclude a small number of outlines values to enhance training efficiency.

\section{Conclusion}

In this paper, we present a novel approach called visual-centric selection approach via agent collaboration, which improve the training efficiency and effectiveness of MLLMs through reduce the noise data. Extensive experiments indicate that , with less than 5\% of the data, models trained with our method achieve comparable or even superior results to state-of-the-art approaches. Our paper demonstration that even a small number of high quality data could surpass a large scale of data. Future work will focus on developing more diverse and comprehensive image quality evaluation metrics to capture a broader range of visual complexities and enhance the robustness of model training.

\section*{Limitations}

Despite our discoveries and improvements, we must acknowledge certain limitations in our work:

\paragraph{Data Size:} Due to resource constraints, our experiments were conducted on a dataset of 200K samples, from which 80K instruction data were finally selected. This limitation restricts our exploration of the application of ViSA on larger datasets. In future work, we aim to extend the use of ViSA to a broader range of open-source datasets.

\paragraph{Theoretical Foundation:} While our method for quantifying image information richness has yielded some promising results, it currently lacks a solid theoretical foundation, limiting the generalizability of our approach. Given the diverse sources and purposes of image data, a comprehensive evaluation of image quality remains a challenging task. We plan to investigate additional theories and methods for image information metrics in future.


\bibliography{custom}

\newpage
\appendix
\label{sec:appendix}

\section{Diversity Perspectives Prompt}
\label{sec:prompt appendix}

We present the default version prompt for diversity perspectives quantification used in our experiment in the~\autoref{tab:full_prompt}. 

\section{Quantification Distribution Analysis}
\label{sec:distrubution appendix}

\begin{figure*}[ht]  
    \centering
    
    \begin{subfigure}[b]{0.45\textwidth}
        \centering
        \includegraphics[width=\linewidth, trim=50 30 50 10, clip]{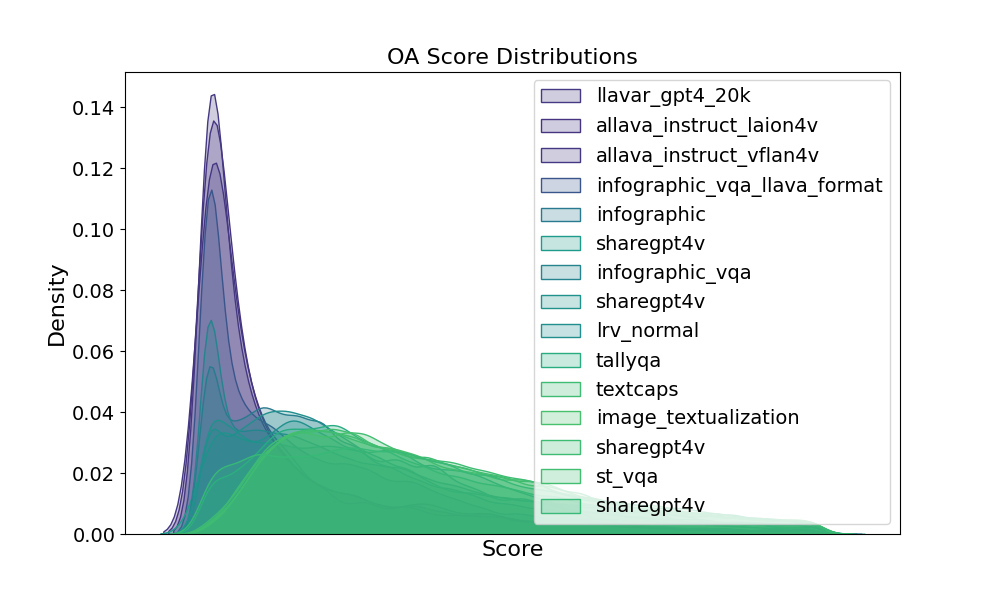}
        \caption{}
    \end{subfigure}
    \hfill
    \begin{subfigure}[b]{0.45\textwidth}
        \centering
        \includegraphics[width=\linewidth, trim=50 30 50 10, clip]{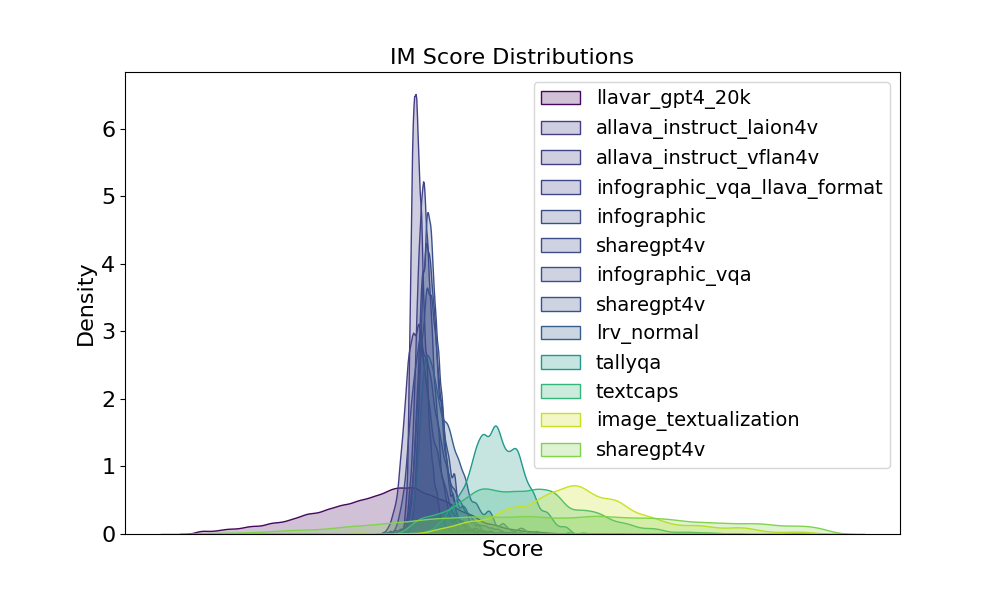}
        \caption{}
    \end{subfigure}
    \hfill
    \caption{Score distributions across different datasets: (a) object alignment, and (b) image-text mutual information.}
    \label{fig:score_distributions_appendix}
\end{figure*}

\paragraph{Object alignment score distribution}
As shown in Figure~\ref{fig:score_distributions_appendix}, our object alignment score  exhibit similar trends with previous segmentation complexity score. However, while the SC score tends to reflect small visual elements such as shapes and color blocks, the OA score is more aligned with visual entities. As a result, the distribution of the OA score is more concentrated. Higher OA scores indicate the presence of more rare and distinctive visual entities, highlighting the uniqueness of the image. In contrast, images with lower OA scores contain more common and less distinctive entities, resulting in lower informational richness. The presence of too many common entities may introduce bias into the visual model’s representation learning, thereby affecting its efficiency.

\paragraph{Image-text mutual information score distribution}
 The distribution of the image-text mutual information score reveals a concerning trend: certain open-source datasets primarily focus on evaluating instruction-following ability, often overlooking the critical relationship between the instruction and the corresponding image. For example, we observed that the llava\_gpt4\_20k dataset frequently contains textual explanations or discussions that are largely unrelated to the image content. Such data not only fails to effectively guide the visual model in learning meaningful visual representations but also introduces significant noise into the visual instruction learning process. This highlights the need for careful curation of datasets to mitigate such mismatches and enhance the quality of visual instruction learning. These findings emphasize the importance of aligning image-instruction pairs in multimodal datasets, ensuring that both components are relevant and complementary to the task at hand. Such principles should guide future efforts in constructing more effective and coherent multimodal datasets.

\section{Case Study}

\begin{figure*}
    \centering
    \includegraphics[width=0.9\textwidth]{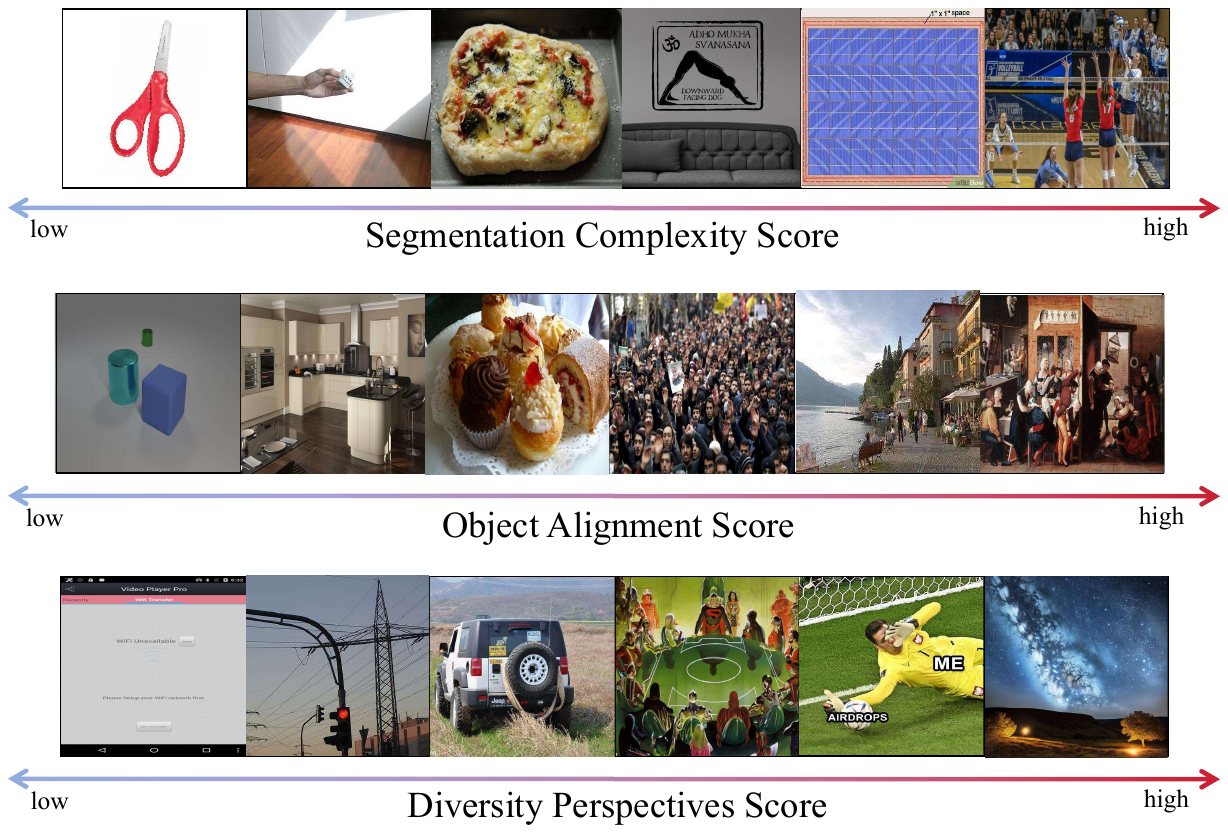}
    \caption{Case study of our proposed image informative metrics. Images on the right exhibit higher scores and richer content, while images on the left correspond to lower scores and simpler content.}  
    \vspace{-4mm}
    \label{fig:case study}
\end{figure*}

To provide a more intuitive demonstration of the proposed image information richness metrics, we present a case study, shown in Figure~\ref{fig:case study}. In this figure, images on the right correspond to higher scores, while images on the left are associated with lower scores. Upon observation, we find that all three proposed metrics effectively filter for images with richer content. The segmentation complexity score favours  simple visual elements, such as shapes and color blocks. For instance, in the SC score examples, the image on the far left—featuring only a single object, a pair of scissors—receives a low SC score, whereas the volleyball match image on the far right, with richer content, achieves a higher score. However, we also present an example of misjudgment: the second image on the right, a fabric with repeated textures, has a relatively high SC score despite its overall simplicity. This highlights the limitation of using basic image segmentation as a sole measure of image complexity.

In contrast, the object alignment score does not suffer from this issue. For example, in the third image on the right, depicting a crowd, the presence of numerous people does not result in an inflated OA score, as the frequent repetition of entities in the image reduces their individual weight. Thus, the image is correctly assigned a higher OA score without any misjudgment.

Regarding the diversity perspectives score, we focus on the multidimensional evaluation of images, which allows us to prioritize "interesting" images. For instance, the third image on the right, a comic, and the second image, a meme, receive high scores for emphasizing specific cultural elements. The photo on the far right earns the highest score due to its aesthetic value, while the screenshot on the far left, primarily composed of text, is assigned a low score due to its lack of visual information.

\section{Evaluation Detail}
\label{sec:evaluation_appendix}

In this section, we provide additional information and details regarding the evaluation benchmarks.
VQAv2 is a widely adopted open-domain visual question answering dataset, consisting of over one million question-answer pairs. For our evaluation, we sample a subset of 10,000 questions. OKVQA is designed to assess models' ability to answer visual questions by leveraging external knowledge sources. Likewise, TextVQA includes over 45,000 text-centric questions based on more than 28,000 images extracted from selected categories of the OpenImages dataset.
To evaluate advanced multimodal reasoning capabilities, MMBench (V1.1) provides a diverse set of 4876 multiple-choice questions compiled from various sources, spanning 20 distinct capability dimensions. MME-RealWorld (MME-RW) serves as a benchmark tailored to real-world applications, focusing on practical scenarios; we utilize its lite version, which consists of 1919 of the most challenging questions. SEED-Bench (SEED) comprises 14232 carefully annotated multiple-choice questions, covering 12 key evaluation dimensions. Lastly, MMMU consists of 1050 challenging questions across a wide range of interdisciplinary tasks, requiring college-level subject knowledge and advanced reasoning skills. For each dataset, we follow the answer-matching strategy recommended in its original paper and compute accuracy with a widely used open-source evaluation toolkit~\citep{DBLP:conf/mm/DuanYQFCLDZZWL024}.

\begin{table*}[ht]
    \centering
    \renewcommand{\arraystretch}{1.1} 
    \scalebox{0.9}{
        \begin{tcolorbox}
                \textbf{The default version of PICA prompt with an example} 
                \tcblower
                \footnotesize
                \# \textbf{Image Complexity Assessment Task} \\
                
                    In this task, you will evaluate the informational richness of an image based on multiple dimensions. Each image may contain multiple layers of information, including various visual elements, contextual settings, emotional expressions, and cultural backgrounds. Please provide a detailed evaluation based on the following rating criteria. \\
                
                \#\# \textbf{Rating Criteria:} \\

                    The overall score will be based on all valid indicators. Images with particularly strong performance in certain dimensions (e.g., complex context or strong emotional expression) will receive higher scores, and weaker performance in other dimensions will not significantly lower the final score. It is recommended to focus on the image’s overall informational layers, element rarity, context building, and uniqueness rather than the absence of a single dimension. \\

                \textbf{1 Point}: Very low information, typically a single subject with a simple background or no clear context. The image lacks depth and multi-dimensional understanding, with few visual elements and no notable uniqueness. \\
                \textbf{2 Point}: Low information, with a few simple elements. It may have basic context or structure but lacks detail and uniqueness. The image is plain and suitable for basic understanding tasks, with weak emotional or contextual expression. \\
                \textbf{3 Point}: Moderate information, with a reasonable number of elements or layers. Context, emotion, or uniqueness are more noticeable, and the image provides moderate multi-dimensional understanding. It has some depth in visual elements and context but doesn’t meet all high-level requirements. \\
                \textbf{4 Point}: High information, with rich elements and clear context, showing uniqueness and diverse layers. The image is detailed, with strong emotional or contextual expression, but may lack some rare or background features. Suitable for in-depth analysis, particularly in context, emotion, culture, or history. \\
                \textbf{5 Point}: Very rich information, with diverse elements, complex scenes, unique perspectives, and abundant detail. The image features rare qualities and deep context, emotion, and cultural background. Ideal for advanced analysis with multi-dimensional depth. \\

                \textbf{\#\# Evaluation Dimensions} \\

                \#\#\# 1. \textbf{Details and Materiality}  \\
                \textbf{Detail Density}: Evaluate the number and complexity of details in the image. The denser the details, the greater the informational content.\\
                \textbf{Material and Texture}: Evaluate whether the image showcases multiple materials (e.g., wood, metal, fabric) and whether the textures are clearly visible. A variety of material layers adds to the visual richness.\\
                \textbf{Detail Layers and Spatial Perception}: Evaluate whether the image presents multiple layers of details, such as clear details in the foreground and a blurred background, creating a sense of depth. \\

                \#\#\# 2. \textbf{Context and Narrative} \\
                \textbf{Context Constructio}: Evaluate whether the image constructs a clear context or scene that conveys rich information through visual elements (e.g., a family gathering, office work, festive events). \\
                \textbf{Narrative Complexity}: Evaluate whether the image depicts one or more complex events or actions, increasing the narrative depth and informational content. \\
                \textbf{Implied Background Story}: Evaluate whether the image implies a potential story or context through visual elements, such as through the environment, actions of people, or objects that hint at social, historical, or cultural contexts. \\

                \#\#\# 3. \textbf{Emotion and Atmosphere} \\
                \textbf{Emotional Expression}: Evaluate whether the image conveys a specific emotion or atmosphere through elements like lighting, color tones, and composition. Emotional expression increases the image's complexity. \\
                \textbf{Emotional Layering}: Evaluate whether the image expresses multiple emotional layers or emotional shifts through different elements. Images with rich emotional layers often have greater depth. \\

                \#\#\# 4. \textbf{Cultural and Historical Context} \\
                \textbf{Cultural Characteristics}: Evaluate whether the image contains elements that reflect a specific cultural, historical, or social context (e.g., distinctive architectural styles, clothing, festivals). \\
                \textbf{Historical Background}: Evaluate whether the image reflects historical events, periods, or characteristics. Historical elements enhance the image’s depth. \\

                \#\#\# 5. \textbf{Camera Angle and Composition} \\
                \textbf{Unique Perspective}: Evaluate whether the image features a unique or creative angle, using uncommon perspectives.\\
                \textbf{Composition Complexity}: Evaluate whether the image’s composition is complex and varied, using techniques like contrast, symmetry, and spatial distribution to enhance the image’s depth. Complex compositions effectively convey more information. \\

                \#\#\# 6. \textbf{Dynamics and Interaction} \\
                \textbf{Dynamic Elements}: Evaluate whether the image includes dynamic elements (e.g., movement of people or objects, hints of time progression). \\
                \textbf{Interactivity}: Evaluate whether  the image depict interaction between elements (e.g., people talking, animals hunting)? Interaction enhances the appeal and complexity of the image. \\

        \end{tcolorbox}
    }
    \caption{The default version of PICA prompt with an example}
    \label{tab:full_prompt}
\end{table*}

\end{document}